# Generating Cyber Threat Intelligence to Discover Potential Security Threats Using Classification and Topic Modeling

**Md Imran Hossen**
School of Computing and Informatics
University of Louisiana at Lafayette
Louisiana, USA

**Ashraful Islam**
School of Computing and Informatics
University of Louisiana at Lafayette
Louisiana, USA

**Farzana Anowar**
Department of Computer Science
University of Regina
Regina, Canada

**Eshtiak Ahmed**
Faculty of IT and Communication Sciences
Tampere University
Tampere, Finland

**Mohammed Masudur Rahman**
Department of Computer Science and Engineering
Bangladesh University of Engineering and Technology
Dhaka, Bangladesh

**Xiali (Sharon) Hei**
School of Computing and Informatics
University of Louisiana at Lafayette
Louisiana, USA

## Abstract

Due to the variety of cyber-attacks or threats, the cybersecurity community enhances the traditional security control mechanisms to an advanced level so that automated tools can encounter potential security threats. Very recently, Cyber Threat Intelligence (CTI) has been presented as one of the proactive and robust mechanisms because of its automated cybersecurity threat prediction. Generally, CTI collects and analyses data from various sources e.g., online security forums, social media where cyber enthusiasts, analysts, even cybercriminals discuss cyber or computer security-related topics and discovers potential threats based on the analysis. As the manual analysis of every such discussion (posts on online platforms) is time-consuming, inefficient, and susceptible to errors, CTI as an automated tool can perform uniquely to detect cyber threats. In this chapter, we identify and explore relevant CTI from hacker forums utilizing different supervised (classification) and unsupervised learning (topic modeling) techniques. To this end, we collect data from a real hacker forum and constructed two datasets: a binary dataset and a multi-class dataset. We then apply several classifiers along with deep neural network-based classifiers and use them on the datasets to compare their performances. We also employ the classifiers on a labeled leaked dataset as our ground truth. We further explore the datasets using unsupervised techniques. For this purpose, we leverage two topic modeling algorithms namely Latent Dirichlet Allocation (LDA) and Non-negative Matrix Factorization (NMF).

***Keywords*** Cyber threat intelligence · CTI · Hacker forum · Cyber-attack · Machine learning · Topic modeling

## 1 Introduction

### 1.1 Background and Motivation

Cybersecurity is one of the most key concerns among users [1, 2, 3]. Cybercriminals are delivering more strong and advanced security threats constantly to gain security control over the networks i.e. Internet [1, 3, 4]. During the previous decades, computer security was bound in only preventing computer viruses and another most common security issue was email spamming [5, 6]. But, in recent years, cybercriminals are making targeted attacks so that they can obtain the whole control from the users. Over the online community, topics related to computer security are the most popular and community members discuss prior and possible security threats in social media, hacker forums [1, 2, 7, 8]. Hacker forums are the most valuable source for computer security related posts, blogs and these posts usually contain vital information about possible computer and cybersecurity threats or holes [1, 2].

Cyber Threat Intelligence, abbreviated as CTI, can be defined as a knowledge and intelligence which assists in collecting, analyzing, understanding, preventing possible cyber-attacks based on gathered data from various security related sources e.g. hacker forums and by transferring analyzed data to possible cyber threats in meaningful representations [2, 3, 8]. According to Gartner [9]:

> *"Threat intelligence is evidence-based knowledge, including context, mechanisms, indicators, implications, and action-oriented advice about an existing or emerging menace or hazard to assets. This intelligence can be used to inform decisions regarding the subject's response to that menace or hazard."*

In general, CTI usually compiles data from diverse sources (usually of unstructured nature) such as the National Vulnerability Database (NVD), social media platforms, e.g., Facebook, Twitter, and computer security forums, and uses scientific methods to convert it to the meaningful representation of threats [2, 3, 10, 11]. In this way, CTI can provide insights into emerging security threats and may help prevent possible and potential cyber-attacks [1, 12].

### 1.2 Problem Statement and Goal

The hackers are constantly posting and sharing information about diverse cybersecurity-related topics on online forums [1, 8]. The posts on these forums can hold data that may help assist in the discovery of cyber threat intelligence. However, manual analysis of this highly unstructured data could be tedious and ineffective [4, 11, 13]. The goal of this work, therefore, is to design and develop a set of procedures to collect, process, and analyze hacker forums data to discover emerging threats. Once we have constructed datasets, classification tasks could be apply to the posts so that classifiers can distinguish security-relevant posts from irrelevant ones. Furthermore, we can categorize security-relevant posts in more specific cyber threat categories e.g. credential leaks, keyloggers, DDoS attacks, etc.

Moreover, we utilize topic modeling algorithms to the whole datasets as well as on each security-relevant class so that we can explore latent topics available in the datasets. For example, if we discover a category that holds a discussion about ransomware, and if our topic modeling algorithm finds keywords like *WannaCry*, and *Petya* for this topic, we would know that these are the current trends in ransomware-based attacks.

The contributions of this research are two-folded.

- Designing and developing a set of procedures to collect, process and analyze data from open source hacker forums.
- Generating CTI from processed data to discover emerging threats by utilizing the knowledge of Information Retrieval (IR). This stage can be done in three phases as following:
    - Identify security-relevant posts from non-relevant/irrelevant ones *(Classification problem)*.
    - Categorize security-relevant posts according to different threat categories *(Classification problem)*.
    - Explore key topics in the dataset using topic modeling algorithms *(Unsupervised learning)*.

## 2 Methodology

This research work is implemented by performing several sequential stages. The following stages are involved in developing our intended goal i.e. mining and classifying hacker forums' data for fathering CTI. Figure 1 demonstrates the steps of the methodology for extracting CTI from the hacker forums.

### 2.1 Data Collection

We first collect data from the posts shared in hacker forums that are posted by computer and network security loving persons, analysts, and experts. Collecting data from hacker forums is a challenging task for several reasons: a) the forums usually restrict the access to their content only to registered users making it difficult to build a web crawler to collect data automatically, b) some forums use CAPTCHAs to prevent automated programs that are used to collect data.

# Generating Cyber Threat Intelligence to Discover Potential Security Threats Using Classification and Topic Modeling

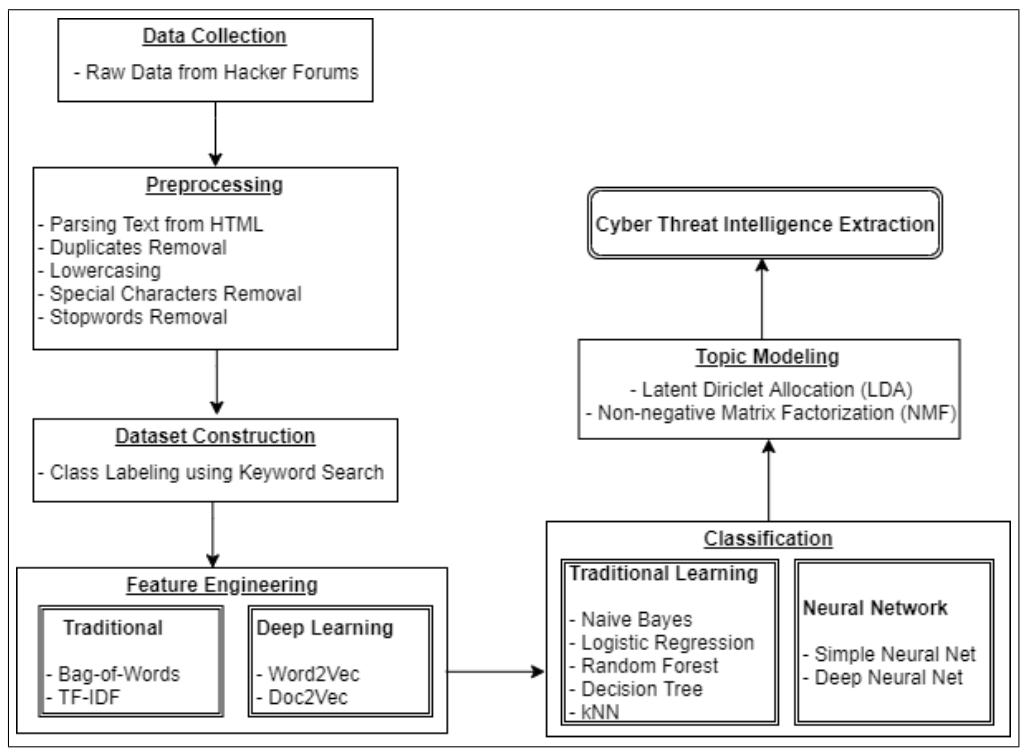

Figure 1: Steps that were followed for extracting CTI.

To make things worse, forums often employ invitation-only registration; one can register only when someone who is already registered on the forum sends him/her an invitation. Alternatively, we may use some publicly available datasets on the internet.

However, we find some decent forums that do not have such restrictions, e.g., [14, 15]. Also, it is noted that we only use the data from the second forum since the first one does not seem to contain a good amount of posts. We build two separate crawlers (using Python) to collect posts from these forums, which in turn will be used to make the corpus. Moreover, we utilized a leaked dataset which is labeled by the experts and it is used as the ground truth dataset for this work from *https://nulled.io*.

## 2.2 Pre-processing and Dataset Construction

Generally, data collected from the web are highly unstructured. Some preprocessing is necessary before we create the final dataset. Hence, first, we clean up the HTML tags and extract the title and content of each post found in the forums, second, we remove stop words, special characters, punctuation after transforming all the text into lower-case. Further, lemmatization is also used so that we can obtain more clean and optimized text collection to build the datasets. Once the data is reasonably cleaned up, we represent each post as a single line document to build a corpus.

Our target is to build two datasets for both binary classification and multi-class classification. We build separate datasets for each security forums we crawled and preprocessed. Both datasets are constructed using the keyword search method where we label each sample based on keyword relevancy.

A list of very common cybersecurity keywords are used for labeling the relevant class are- *adware, antivirus, botnet, backdoor, crack, crimeware, crypter, ddos, downloader, dropper, exploit, firewall, hijack, infect, keylogger, malware, password, ransomware, reverse, shell, rootkit, scanner, shell, code, security, spam, spoof, spyware, trojan virus, vulnerability, worm, zero-day, stealware*.

We attempt the following approaches for building both the datasets.

### 2.2.1 Binary Dataset Construction

Binary dataset has two labels or classes i.e. relevant, irrelevant. The samples bearing the posts containing keywords relevant to computer security are labeled as relevant. Some common relevant keywords that we approach are 'exploit', 'keylogger', 'reverse shell', 'antivirus', 'ransomware', etc. Other samples are labeled as irrelevant as they contain irrelevant keywords such as 'song', 'tv show', 'browsing', 'football', etc. Since the number of security keywords are not comprehensive we label a post as irrelevant if it satisfies these two properties –

- None of the keywords from the relevant class is present, and
- Any computer security related keywords are absent.

Table 1 shows two samples from the binary dataset for better understanding where relevant keywords are underlined. Figure 2 presents the distribution of sampler for each class i.e. relevant, irrelevant. Table 1 contains examples of posts labeled in the binary dataset.

Table 1: Two samples collected from binary dataset (Posts collected from various sources found in Internet for dataset construction).

| Post | Class Label |
|---|---|
| "... Here is a .php file from an exploit, downloaded from ... Apache APR is prone to a vulnerability that may allow attackers to cause a denial-of-service condition ..." | Relevant |
| "... Good work, its good to learn how to do things yourself. Its a very rewarding experience ..." | Irrelevant |

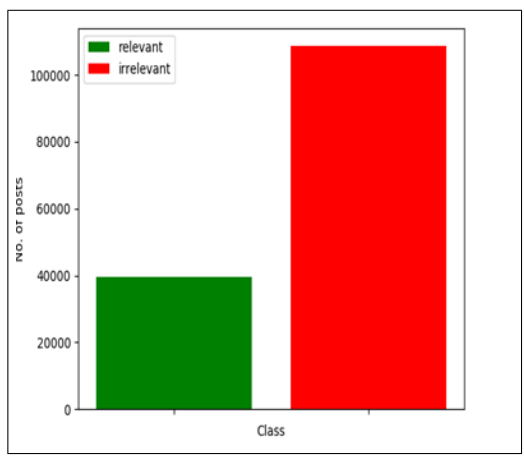

Figure 2: Samples (posts) distribution per class i.e. relevant, irrelevant

### 2.2.2 Multi-class Dataset Construction

At this phase, we build another dataset from the samples labeled as relevant in the binary dataset. This dataset contains samples labeled in six categories and the categories are-

- Credential Leaks (Keywords: username, password, pass list, etc.)
- Keylogger (Keyword: keylogger )
- DDoS Attack (Keywords: ddos, denial of service, server, etc.)
- Remote Access Trojans (Keywords: rat, trojan, remote access, etc.)
- Cyrpters (Keywords: crypter, fud, etc.)
- SQL Injection (Keywords: sql, injection, id=, etc.)

Table 2 contains examples of posts labeled in the multi-class dataset.

Generating Cyber Threat Intelligence to Discover Potential Security Threats Using Classification and Topic Modeling

Table 2: Few samples collected from the multi-class dataset constructed from the relevant class (Posts collected from various sources found in Internet for dataset construction).

| Post | Class Label |
|---|---|
| "Leaked data for *** <br> USERNAME / PASSWORD <br> hope0507 : *** <br> jimsam100234:***" | Credential Leaks |
| "The Best Keylogger logs all keystrokes, mouse clicks, applications ... sent to you via mail so you can monitor your computer without being there" | Keylogger |
| "If you're not interested in creating a botnet, with a RAT for example ... Which use servers to DDoS targets ... Written on mobile-phone, sorry for the formatting" | DDoS attack |
| "Initial checker <br> $newHost = $host.'/interface/ipsconnect.php'; <br> $sql = 'SELECT COUNT(*) FROM members'; <br> ... $response = SendPost($newHost, $data)"; | SQL Injection |
| "Plasma HTTP BotNet allows you to remotely control computers, known as Clients ... Persistent (Miner is re-injected if terminated) (...truncated)" | Crypter |
| "Application name: Imminent Monitor Application description: Advanced RAT (The others cracked/leaked here ... Vendor's URL: http://www.hackforums.net/showthread.php? /" | Remote Access Trojans |

### 2.3 Feature Engineering

Machine learning algorithms cannot directly work with textual data [1, 12, 16, 17]. We need to vectorize the documents before feeding them as inputs to any learning algorithms, either supervised or unsupervised [13, 18]. To this end, we explore the best of both worlds for this purpose. First, we use standard feature engineering techniques such as bag-of-words (BOW) and term frequency-inverse document frequency (TF-IDF)-based weights. Further, both binary term weights and term-frequency weights measurements are used for the BOW model. Moreover, for TF-IDF based weighting we use two varieties- word-level unigram and word-level bigram features.

Second, we conduct the experiment with two contemporary and more advanced feature engineering techniques based on deep learning namely, *Word2Vec* and *Doc2Vec* models. *Word2Vec* is a word-embedding model that learns the semantic between words in the corpus [19]. There are several pre-trained *Word2Vec* models available on the internet. *Doc2Vec* model performs very well while using with sentiment analysis data [20, 21, 22]. However, none of them are trained on the cybersecurity-relevant corpus. For this reason, we do not use any trained models and train a *Word2Vec* model (with feature_size=100) on our own dataset. Since the *Word2Vec* model by default does not provide document-level vectors, we obtain a 100-dimensional vector for each word in a document and summed them up and calculate the final averaged document vector. We do the same for all the documents in the dataset. We also train a *Doc2Vec* model on our dataset which directly learns and provides the document-level vectors [23].

### 2.4 Supervised Method: Classification

We first carry out classification tasks to see how accurately cybersecurity-relevant posts can be separated and identified from the non-security posts. This is an important step since the majority of the posts in the hacker forum has nothing to do with security and serve as noises. Filtering out the noises, provides us the chance to focus more on what we actually care about and that is security-relevant posts. It also allows us to further inspect and investigate the security-relevant posts for potential cyber threats. We also evaluate the performance of classification for our multiclass dataset.

In our experiment, we use five different classification algorithms along with two neural network learning classifiers. The classifiers that we use are: Naive Bayes (NB), Logistic regression (LR), Random Forest (RF), Decision Tree (DT) and k-nearest neighbors (kNN). We also use one shallow neural network and one deep neural network-based classifiers.

### 2.5 Unsupervised Method: Topic Modeling

Topic modeling is the process of identifying and determining topics in a set of documents. It is an unsupervised learning approach and can be a powerful tool to find possible latent topics (in the form of keywords) in a large unlabeled dataset [24, 25]. For instance, say we have a dataset that contains newspaper articles from some major online news agencies. We provide the number of topics to the algorithm, it will provide top *k* (number of topics) per topic. By looking at the keywords for each topic we may be able to tag topics like politics, technology, culture, and so on. It can also be useful for a labeled dataset to further explore key topics in each class. We use two popular commonly used topic modeling algorithms in this work. The algorithms are: 1) Latent Dirichlet allocation (LDA), and 2) Non-negative Matrix Factorization (NMF). We consider 'frequency weights' as a feature to LDA as we find TF-IDF weights do not seem to make any difference for this algorithm. In contrast, TF-IDF weights based features are used for the NMF algorithm.

## 3 Experimental Setup

Data that is used for this work, is collected on a server with 12 Intel R Xeon R E5-2667 CPUs, and 64 GB of RAM running Debian Linux operating system with Linux kernel version 4.18. We develop and test our code for classification on a server with 6 Intel R Xeon R E5-2667 CPUs, an NVIDIA GeForce RTX 2070 GPU, and 96GB of RAM running the Arch Linux operating system.

We process the data using standard regular expressions. However, for lemmatization, we use the spaCy library. We also utilize several open-source Python libraries throughout this research work. We use the scikit-learn libraries to develop the classifiers. The neural network-based classifiers are coded in Keras. We use both gensim and scikit-learn libraries for topic modeling. The gensim library is also used for training *Word2Vec* and *Doc2Vec* models.

## 4 Experimental Results

In the first phase of our experiment, we utilize five regular and two neural network-based classifiers. In all of our experiments, we split the whole data for organizing datasets for both training (67%) and testing (33%). We validate our code on two datasets, IMDB movie review dataset (binary) [26] and twenty newsgroups dataset (multiclass) [27]. We find that our results are comparable with the given benchmark results. Specifically, the accuracies vary from 76% to 88% for different classifiers with different feature sets on IMDB movie review datasets. For the twenty newsgroups dataset (only 8 classes have been considered) the accuracies vary from 79% to 89%.

Table 3 lists the performance metrics based on accuracy for different classifiers for *forums.hak5.org* data sets. Table 4 presents the performance metrics in accuracy for different classifiers for the *nulled.io* data sets or our ground truth data sets.

For topic modeling we utilize two popular algorithms: a) Latent Dirichlet Allocation (LDA) and b) Non-negative Matrix Factorization (NMF). These algorithms require some *k* (number of topics) to be passed as an input. Finding the optimal value of *k* requires some experimentation. But our goal is about determining the best value of *k*, rather we want to explore the datasets and get some more insights of the datasets. We use *k* as 10 for binary datasets and apply the algorithm as a dataset as a whole. For multinomial datasets we run LDA and NMF separately on individual categories. In any cases, we show only the top 05 keywords per category. The topic modeling results are listed in Tables 5-8 (showing only 05 topics in each table and 'Credentials' category for the simplicity of the paper).

Table 3: Classification accuracy (%) obtained by classifiers on the *forums.hak5.org* dataset

| Dataset | Classifier | Feature Extraction Method | | | | | |
|---|---|---|---|---|---|---|---|
| | | *BOW* (Binary term weights) | *BOW* (TF weights) | *TF-IDF* (Unigram) | *TF-IDF* (Bigram) | *Word2Vec* | *Doc2Vec* |
| **Binary** | Naive Bayes | 79.96 | 78.96 | 75.36 | 64.65 | - | - |
| | Logistic Regression | 93.67 | 92.75 | 90.69 | 68.87 | 64.64 | 73.85 |
| | Random Forest | 89.88 | 89.89 | 88.3 | 68.55 | 58.4 | 62.88 |
| | Decision Tree | 93.4 | 92.8 | 92.18 | 67.03 | 56.5 | 58.49 |
| | kNN | 47.42 | 57.08 | 60.01 | 53.29 | 59.96 | 62.61 |
| **Multiclass** | Naive Bayes | 85.67 | 83.06 | 70.68 | 68.4 | - | - |
| | Logistic Regression | 97.39 | 96.09 | 83.06 | 75.24 | 48.86 | 94.46 |
| | Random Forest | 89.25 | 94.14 | 91.86 | 70.03 | 48.53 | 63.19 |
| | Decision Tree | 97.39 | 97.07 | 97.07 | 64.5 | 38.76 | 54.72 |
| | kNN | 88.27 | 85.99 | 77.2 | 28.66 | 49.19 | 73.62 |
| | Simple Neural Net | 86.97 | 85.34 | 83.81 | 78.83 | - | - |
| | Deep Neural Net | 77.85 | 87.3 | 79.48 | 77.85 | - | - |

## 5 Discussion on Results

In almost all cases the kNN algorithm yields the worst performance comparing to other classifiers. It is interesting to note that the simple term-frequency weights as feature provide better (or similar) results than TF-IDF. We also found

# Generating Cyber Threat Intelligence to Discover Potential Security Threats Using Classification and Topic Modeling

Table 4: Classification accuracy (%) obtained by classifiers on the *nulled.io* dataset

| Dataset | Classifier | BOW (Binary term weights) | BOW (TF weights) | TF-IDF (Unigram) | TF-IDF (Bigram) | Word2Vec | Doc2Vec |
|---|---|---|---|---|---|---|---|
| Binary | *Naive Bayes* | 91.65 | 90.75 | 91.61 | 90.71 | - | - |
| | Logistic Regression | 93.26 | 93.06 | 92.01 | 93.71 | 93.1 | 92.85 |
| | Random Forest | 92.96 | 93.16 | 91.76 | 9067 | 9.18 | 92.5 |
| | Decision Tree | 94.62 | 94.77 | 92.75 | 88.73 | 87.41 | 85.26 |
| | kNN | 47.56 | 61.44 | 78.95 | 90.42 | 92.95 | 92.8 |
| Multiclass | Naive Bayes | 74.94 | 75.66 | 70.01 | 72.53 | - | - |
| | Logistic Regression | 87.06 | 86.29 | 87.59 | 57.95 | 42.49 | 88.31 |
| | Random Forest | 84.21 | 85.27 | 85.9 | 76.2 | 41.04 | 64.8 |
| | Decision Tree | 86.96 | 86.96 | 87.01 | 72.86 | 34.67 | 52.34 |
| | kNN | 67.84 | 69.34 | 74.17 | 46.45 | 38.88 | 77.98 |
| | Simple Neural Net | 86.53 | 86.05 | 85.22 | 77.84 | - | - |
| | Deep Neural Net | 87.11 | 86.53 | 85.66 | 77.64 | - | - |

Table 5: Result after applying topic modeling on *forums.hak5.org* Binary dataset.

**Dataset: forums.hak5.org BINARY**

**Topic Modeling Method: LDA**

| Topic | Keywords |
|---|---|
| 0 | ['password' 'see' 'google' 'know' 'site' 'android' 'github' 'user' 'use' 'windows'] |
| 1 | ['script' 'windows' 'file' 'usb' 'payload' 'files' 'work' 'ducky' 'run' 'drive'] |
| 2 | ['pineapple' 'connect' 'wifi' 'internet' 'get' 'connection' 'network' 'port' 'router' 'connected'] |
| 3 | ['email' 'antenna' 'mail' 'darren' 'signal' 'spam' 'cp' 'speed' 'wait' 'antennas'] |
| 4 | ['http' 'www' 'php' 'hak5' 'https' 'forums' 'code' 'like' 'index' 'page'] |

**Topic Modeling Method: NMF**

| Topic | Keywords |
|---|---|
| 0 | ['like' 'get' 'know' 'use' 'something' 'work' 'time' 'want' 'see' 'really'] |
| 1 | ['pineapple' 'wifi' 'internet' '172' 'mark' 'connect' 'firmware' 'via' 'connected' 'connection'] |
| 2 | ['php' 'hak5' 'index' 'forums' 'topic' 'https' 'showtopic' 'http' 'forum' 'thread'] |
| 3 | ['usb' 'power' 'drive' 'ducky' 'hub' 'rubber' 'battery' 'adapter' 'port' 'device'] |
| 4 | ['file' 'script' 'payload' 'ducky' 'payloads' 'bunny' 'run' 'exe' 'files' 'command'] |

that word-level unigram and bigram for TF-IDF weighting have a minor impact on final performance metrics. Further, using *Word2Vec*-based feature engineering techniques do not provide any significant performance boost. In some instances, specifically for multiclass datasets, they actually seem to generate much less accurate results comparing to TF and TF-IDF weights. One reason can be the fact that we only use 100-dimensional vectors to represent the documents for both *Word2Vec* and *Doc2Vec*. Besides, the multiclass dataset has more categories and a small number of samples per class when comparing to the binary-class datasets. Increasing the size of the feature vector may result in enhanced performance. However, it will consume significant computing resources to train such a model. For this reason, we refrain from doing so. We can also see that *Doc2Vec*-based models perform remarkably better than *Word2Vec* models. The *Doc2Vec* model learns word-level and document level embedding simultaneously. Consequently, it outperforms *Word2Vec* where we use simple word vector averaging technique to derive document-level vectors.

It is also noticeable that none of the neural network-based classifiers (shallow neural network and deep neural network), is unable to beat traditional classifiers we use in this experiment. While neural networks algorithms, especially deep neural networks, have made a breakthrough in several computer vision applications and outperform all machine learning-based approaches, their performance for text classification is about the same as the regular classifiers. Using neural network algorithms specialized in Natural Language Processing (NLP) such as Recurrent Neural Network (RNN), and long short-term memory (LSTM), perform better in this case [20]. However, we did not use them. There are two reasons for this decision. First, training RNN based models requires a good amount of computing resources and they are very slow. Second, our basic classifiers have already provided with promising results.

Table 6: Result after applying topic modeling on *forums.hak5.org* Multiclass dataset.

| Dataset: forums.hak5.org MULTI-CLASS | |
|---|---|
| **Category: 'Credentials'** | |
| **Topic Modeling Method: LDA** | |
| Topic | Keywords |
| 0 | ['password' 'username' 'get' 'login' 'use' 'like' 'user' 'know' 'account' 'need'] |
| 1 | ['password' 'username' 'file' 'pineapple' 'root' 'http' 'use' 'using' 'get' 'ssh'] |
| 2 | ['hashcat' 'use' 'john' 'using' 'txt' 'net' 'post' 'username' 'mimikatz' 'gmail'] |
| 3 | ['barcode' 'assets' 'date' 'ass_tag' 'echo' 'strlen' 'mysql_query' 'die' 'simpleassets' 'update'] |
| 4 | ['network' 'wan2' 'option' 'uci' 'set' '172' 'etc' 'interface' 'config' 'device'] |
| **Topic Modeling Method: NMF** | |
| Topic | Keywords |
| 0 | ['password' 'root' 'username' 'user' 'login' 'file' 'use' 'get' 'account' 'server'] |
| 1 | ['wan2' 'uci' 'network' 'set' '12d1' 'iptables' 'firewall' 'sleep' '172' 'init'] |
| 2 | ['wifi' 'pineapple' 'connect' 'network' 'wpa2' 'enterprise' 'connected' 'internet' 'client' 'using'] |
| 3 | ['php' 'splash' 'nodogsplash' 'portal' 'auth' 'page' 'html' 'evil' 'call' 'www'] |
| 4 | ['option' 'config' '255' 'ifname' 'proto' 'interface' 'device' 'ht_capab' 'netmask' 'ipaddr'] |

Table 7: Result after applying topic modeling on *nulled.io* binary dataset.

| Dataset: nulled.io BINARY | |
|---|---|
| **Topic Modeling Method: LDA** | |
| Topic | Keywords |
| 0 | ['http' 'www' 'color' 'https' 'application' 'php' 'url' 'topic' 'hack' 'nulled'] |
| 1 | ['bol' 'bot' 'scripts' 'legends' 'nulled' 'download' 'work' 'login' 'use' 'auth'] |
| 2 | ['php' 'gmail' 'hotmail' 'net' 'yahoo' 'hide' 'http' 'index' 'html' 'server'] |
| 3 | ['hide' 'https' 'file' 'www' 'http' 'download' 'spoiler' 'virus' 'link' 'virustotal'] |
| 4 | ['local' '255' 'end' 'function' 'class' 'user' 'bot' 'serverversion' 'autoupdatermsg' 'true'] |
| **Topic Modeling Method: NMF** | |
| Topic | Keywords |
| 0 | ['like' 'script' 'get' 'use' 'know' 'thanks' 'bot' 'crack' 'working' 'help'] |
| 1 | ['legends' 'bot' 'dominate' 'login' 'download' 'available' 'enemies' 'occur' 'monitoring' 'freeze'] |
| 2 | ['hide' 'http' 'mega' 'password' 'https' 'enjoy' 'download' 'upvote' 'html' 'mediafire'] |
| 3 | ['bol' 'scripts' 'vip' 'studio' 'script' 'exe' 'bolvip' 'hijacker' 'cracked' 'use'] |
| 4 | ['nulled' 'www' 'https' 'auth' 'topic' 'php' 'http' 'access' 'need' 'cracked'] |

Our experimental results show that running topic modeling algorithms on binary dataset does not uncover security-relevant keywords. However, when we apply these algorithms ((LDA and NMF) on multi-class datasets as a whole and separately on each security-relevant category, they provide us with topics and keywords that make much sense in the context of their security relevance. Further, we find that the quality of generated keywords for the given topics depends significantly on preprocessing. If we do not preprocess the data properly, the algorithm may get lost in clutter of words such as stop words and other commonly used words, and the final outcome may not represent the latent topics in the dataset.

## 6  Challenges and Future Scopes

We face several issues throughout this research work, for example, the first major challenge is the scarcity of open source threat intelligence (OSTI) datasets for either analysis or benchmarking. We finally manage to get a leaked dataset for *nulled.io* forum from the internet archive website, *https://archive.org* that we use as a ground truth. The second challenge is to collect data from hacker forums for actual experimentation. We observe that most hacker forums employ some kind of anti-crawling technology, such as CAPTCHA, making it difficult to collect data using an automated program like web crawler. Further, some forums only allow invite-only registrations. It means we cannot get access to the forum unless we have an account in it and to get an account we need to get an invitation from a registered member of that forum first. We, however, managed to collect data from a popular forum name *forums.hak5.org* since it does not have such restrictions.

Table 8: Result after applying topic modeling on nulled.io Multiclass dataset.

| Dataset: nulled.io MULTI-CLASS | |
|---|---|
| **Category: 'Credentials'** | |
| **Topic Modeling Method: LDA** | |
| Topic | Keywords |
| 0 | ['username' 'password' 'file' 'download' 'email' 'level' 'hide' 'region' 'pages' 'summoner'] |
| 1 | ['user' 'password' 'username' 'php' 'http' 'admin' 'login' 'hide' 'www' 'script'] |
| 2 | ['var' 'steam' 'system' 'string' 'using' 'function' 'require' 'new' 'windows' 'data'] |
| 3 | ['user' 'action' 'password' 'email' 'login' 'https' 'host' 'username' 'agent' 'form'] |
| 4 | ['password' 'username' 'account' 'accounts' 'get' 'email' 'use' 'like' 'need' 'login'] |
| **Topic Modeling Method: NMF** | |
| Topic | Keywords |
| 0 | ['action' 'user' 'email' 'agent' 'login' 'host' 'form' 'https' 'referer' 'http'] |
| 1 | ['account' 'password' 'username' 'hide' 'bol' 'accounts' 'get' 'login' 'nulled' 'use'] |
| 2 | ['checker' 'check' 'delay' 'accounts' 'killerabgg' 'owned' 'pvpnetconnect' 'library' 'last' 'current'] |
| 3 | ['rune' 'pages' 'rank' 'validated' 'skins' 'champions' 'level' 'unranked' 'summoner' 'region'] |
| 4 | ['user' 'export' 'scrape' 'press' 'select' 'pick' 'number' 'page' 'list' 'check'] |

One limitation of our work is that we construct the dataset using a simple keyword searching technique. This process may result in biased measurements. A possible future extension of this work would be using more advanced dataset construction purposes. Alternatively, it is possible to refine and clean the relevant and non-relevant classes using the relevance feedback provided by some human experts in the cybersecurity domain.

## 7 Conclusion

Our study collected and analyzed data from a popular hacker forum, forums.hak5.org, for the purpose of identifying and classifying possible cyber threat intelligence. We used both standard and state-of-art deep-learning techniques for feature extraction and document vectorization purposes. We then developed and utilized several machine learning and neural network models for classifications. Our results showed that it is possible to classify security-relevant posts and irrelevant one with high accuracy. We further evaluate the performance of these algorithms on the multiclass dataset. Moreover, we utilized two topic modeling algorithms as an unsupervised learning approach. We were able to extract the top words for a different number of topics on these datasets using LDA and NMF topic modeling algorithms. Further, we applied all these approaches to a labeled dataset named *nulled.io*.

[28]